\documentclass[fleqn,10pt,twocolumn]{SICE_FES25}
\usepackage{amsmath}
\usepackage{booktabs}
\usepackage{fancyhdr}
\usepackage{xcolor}

\title{Assessing the Value of Visual Input: A Benchmark of Multimodal Large Language Models for Robotic Path Planning}

\author{Jacinto Colan${}^{1\dagger}$, Ana Davila${}^{2}$ and Yasuhisa Hasegawa${}^{2}$}
\speaker{Jacinto Colan}

\affils{${}^{1}$Department of Micro-Nano Mechanical Science and Engineering, Nagoya University, Nagoya, Aichi, Japan\\
${}^{2}$Institutes of Innovation for Future Society, Nagoya University, Nagoya, Aichi, Japan\\
}
\abstract{%
Large Language Models (LLMs) show potential for enhancing robotic path planning. This paper assesses visual input's utility for multimodal LLMs in such tasks via a comprehensive benchmark. We evaluated 15 multimodal LLMs on generating valid and optimal paths in 2D grid environments, simulating simplified robotic planning, comparing text-only versus text-plus-visual inputs across varying model sizes and grid complexities. Our results indicate moderate success rates on simpler small grids, where visual input or few-shot text prompting offered some benefits. However, performance significantly degraded on larger grids, highlighting a scalability challenge. While larger models generally achieved higher average success, the visual modality was not universally dominant over well-structured text for these multimodal systems, and successful paths on simpler grids were generally of high quality. These results indicate current limitations in robust spatial reasoning, constraint adherence, and scalable multimodal integration, identifying areas for future LLM development in robotic path planning.}

\fancypagestyle{firstpage}{
  \fancyhf{}
  
  \fancyhead[L]{
    \vspace*{-25pt} 
    \parbox{\textwidth}{
      \textcolor{blue}{\small This work has been accepted at the 2025 SICE Festival with Annual Conference (SICE FES) and submitted to the IEEE for possible publication. Copyright may be transferred without notice, after which this version may no longer be accessible.}
    }
  }
}

\begin{document}

\maketitle
\thispagestyle{firstpage}


\section{Introduction}

Effective path planning is fundamental to autonomous robotic systems, enabling navigation through complex and dynamic environments. While traditional algorithms provide established solutions, the integration of semantic understanding and commonsense reasoning, as potentially offered by Large Language Models (LLMs), presents a developing area for enhancing the intuitiveness and adaptability of robot planning. With the evolution of these models, particularly multimodal LLMs capable of processing both textual and visual information, their application to robotics tasks involving spatial layout comprehension and constraint adherence is of growing interest. A systematic, comparative evaluation is necessary to understand the performance characteristics of these models, especially regarding the contribution of visual data when used with well-structured textual descriptions for planning.

This paper examines the role of visual input for multimodal LLMs within the context of robotic path planning. Although real-world robotic navigation contends with continuous state spaces and sensor uncertainties, foundational spatial reasoning and constraint adherence can be systematically benchmarked in structured settings. We utilize 2D grid-based path generation scenarios as a controlled testbed to simulate simplified robotic planning tasks. This approach allows for a focused assessment of an LLM's ability to generate valid and optimal trajectories from a designated start to a goal position while avoiding obstacles. Such scenarios, while abstract, encapsulate core elements of spatial understanding and sequential decision-making relevant to robotic autonomy.

We present a comprehensive benchmark assessing a diverse set of inherently multimodal LLMs. Their performance is evaluated when provided with only textual descriptions of the environment and when they receive both textual information and a corresponding visual representation of the grid. By systematically comparing success rates, path optimality, and the impact of problem complexity (grid size), this work aims to investigate the utility and current limitations of leveraging visual input alongside text for spatial planning tasks performed by contemporary multimodal LLMs. The findings are intended to offer insights relevant to their potential deployment in robotic systems.

\section{Related works}

The integration of Large Language Models (LLMs) into robotic systems has enabled new approaches to interpret high-level human instructions and generate task plans \cite{ref1_old,ref2_old,ref1}. Early work focused on text-based LLMs \cite{ref2}, with notable approaches such as SayCan \cite{ref3} that ground language commands in robot capabilities. Other methods translated natural language goals into formal planning languages (e.g., PDDL) or code policies, sometimes using structured prompts such as ProgPrompt \cite{ref4} to improve reliability. A key challenge in this domain is to ensure that instructions are unambiguous and effectively translated into actions, whether through voice interfaces \cite{ref5} or by using LLMs themselves to detect and resolve ambiguity in commands \cite{ref6}. These systems often separate high-level planning from low-level execution \cite{ref3,ref4}, a concept also explored in complex, non-LLM-based hybrid planning frameworks \cite{ref7}. However, text-only models can lack physical grounding and face challenges with spatial reasoning without perceptual input \cite{ref2_old,ref3}, which has led to an increased focus on multimodal approaches.

To address the grounding challenge, Multimodal Large Language Models (MLLMs), were developed, integrating visual perception with linguistic reasoning \cite{ref8,ref7_old}. This has contributed to the development of more end-to-end systems unifying perception, planning, and action. Models like PaLM-E \cite{ref9} demonstrated injecting sensor data into the LLM embedding space for embodied tasks, while VLAs like RT-2 \cite{ref1_old0} represented robot actions as output tokens and were reported to show emergent reasoning capabilities. Concurrently, methods like VIMA \cite{ref1_old1} used multimodal prompting for manipulation tasks. These approaches generally operate on the assumption that vision provides necessary grounding for embodied tasks \cite{ref8}, an assumption examined in this paper.

LLMs and MLLMs have been specifically applied to robotic path planning. Hybrid approaches combine functionalities of LLMs with classical algorithms, such as LLM-A* \cite{ref1_old2} using LLM heuristics to guide A* search, or LLM-Advisor \cite{ref1_old3} suggesting modifications to paths from other planners. Instruction-based navigation frameworks like LLM-Planner/DCIP \cite{ref1_old4} integrate LLMs with occupancy grids, while Guide-LLM \cite{ref1_old5} uses text-based topological maps for indoor assistance. The environmental representation provided—textual descriptions, grid coordinates \cite{ref1_old6}, topological maps \cite{ref1_old5}, or direct visual input \cite{ref9,ref1_old4}—can significantly influence spatial reasoning performance \cite{ref1_old6}.

Despite the recognized potential of visual input, MLLMs encounter challenges in spatial reasoning, including relational and transformational understanding \cite{ref1_old7}. In some instances, visual input can also impair performance (``cross-modal distraction") \cite{ref1_old8}, and processing visual data adds computational cost \cite{ref1_old0}. While benchmarks exist for LLM planning (e.g., PPNL for text-based grid navigation \cite{ref1_old9}) and general embodied AI (e.g., EmbodiedBench \cite{ref2_old0}), a dedicated benchmark assessing the specific value of visual input for MLLM-based robotic path planning is less established. This paper aims to address this gap by benchmarking the contribution of visual input for MLLMs in robotic path planning.

\section{Methodology}
\label{sec:methodology}

This research employs a quantitative benchmarking approach to evaluate the spatial reasoning and planning capabilities of various Large Language Models (LLMs) on grid-based navigation tasks. The core task for the LLMs is to generate a complete, valid trajectory from a specified start to a goal position in a single request, given the grid's configuration including explicit rules, and, for multimodal models, a visual representation of the grid.

\subsection{Grid Environment and Problem Generation}
We define a 2D grid environment where each cell can be empty, an obstacle (termed `bomb'), a start position (`S'), or a goal position (`G'). A suite of grid problems was procedurally generated with varying sizes (e.g., $8 \times 8$, $20 \times 20$) and bomb densities, based on a defined procedural generation algorithm. For each generated problem instance:
\begin{itemize}
    \item The start (e.g., top-left) and goal (e.g., bottom-right) positions are typically fixed for a given size, though the generation logic can support randomized placements before obstacles are assigned.
    \item Obstacle locations are randomly assigned within the grid, excluding the start and goal positions. The generation process ensures that a solvable path from `S' to `G' exists.
    \item The optimal path length (shortest number of steps from `S' to `G' avoiding obstacles) is pre-calculated using a standard Breadth-First Search (BFS) algorithm.
    \item Each problem instance, detailing grid dimensions, the coordinates of `S', `G', all obstacle locations, and the optimal path length, is stored as a structured JSON file.
    \item A corresponding $512 \times 512$ pixel PNG image representation of the grid is also generated. For prompting multimodal LLMs, this image displays the `S' and `G' locations. Traversable cells are depicted as white, while other cells (implicitly obstacles) are non-white. Obstacle locations are not explicitly marked with a distinct symbol (e.g., `B') in the image provided to the LLM, encouraging reliance on visual interpretation of obstacles or correlation with textual information. This visual representation is generated such that obstacles appear as non-white cells.
\end{itemize}
The evaluation utilizes a fixed set of 20 pre-generated grid problems per grid size, loaded from a designated directory of problem instances, to ensure consistency across all model evaluations.

\subsection{Models Evaluated}
We selected a diverse set of multimodal state-of-the-art LLMs, as detailed in Table \ref{tab:model_details}. Model selection aimed to cover a range of model families, sizes (Small, Medium, Large, including Mixture of Experts (MoE) architectures), and access methods (open-weights or proprietary API). All models were accessed via a unified API endpoint, using a configured API key.

\begin{table}
\caption{Details of LLMs Evaluated in the Benchmark.}
\label{tab:model_details}
\begin{center}
\resizebox{0.48\textwidth}{!}{
\begin{tabular}{|l|l|l|l|}
\hline
\textbf{Model Name} & \textbf{Developer} & \textbf{Size Category} & \textbf{Access} \\ \hline
Llama 3.2 & Meta & Small & Open \\ \hline
Gemma 3 & Google DeepMind & Small & Open \\ \hline
GPT-4.1 Nano & OpenAI & Small & Proprietary \\ \hline
Mistral Small & Mistral AI & Medium & Open \\ \hline
Gemma 3 & Google DeepMind & Medium & Open \\ \hline
Claude 3.5 Haiku & Anthropic & Medium & Proprietary \\ \hline
GPT-4.1 Mini & OpenAI & Medium & Proprietary \\ \hline
GPT-4.1 & OpenAI & Large & Proprietary \\ \hline
GPT-4o & OpenAI & Large & Proprietary \\ \hline
Claude 3.7 Sonnet & Anthropic & Large & Proprietary \\ \hline
Llama 4 Maverick & Meta & Large (MoE) & Open \\ \hline
Llama 4 Scout & Meta & Large (MoE) & Open \\ \hline
DeepSeek V3 & DeepSeek & Large & Open \\ \hline
Mistral Large 2 & Mistral AI & Large & Proprietary \\ \hline
Nova Pro & Amazon & Large & Proprietary \\ \hline
\end{tabular}%
}
\end{center}
\footnotesize{Size Category: Small, Medium, Large. MoE indicates Mixture of Experts architecture. Parameter counts are generalized. Access: ``Proprietary" indicates access via a proprietary API. ``Open" indicates availability through open weights or an open-source license.}
\end{table}

\subsection{Experimental Procedures and Prompting Strategies}
All LLMs were tasked with generating a full trajectory as a Python list of coordinate tuples (e.g., $[(\text{row1}, \text{col1}), (\text{row2}, \text{col2}), \dots)])$ in a single API call. The generated paths are constrained by the rule that cells cannot be revisited, a constraint strictly enforced during path validation. A low temperature parameter ($0.1$) was used for LLM queries to promote deterministic and consistent outputs. The LLM responses, expected to be a list of coordinates, were parsed using a dedicated extraction routine.

\subsubsection{Experiment 1: Text-Only LLM Evaluation}
Text-only LLMs received the grid problem information through a structured textual description. This description, systematically generated for each problem instance, included:
\begin{itemize}
    \item Grid dimensions (e.g., ``Grid Size: 5x5 (coordinates from (0,0) to (4,4))").
    \item Start position  (e.g., ``Start Position (S): (1,1)").
    \item Goal position  (e.g., ``Goal Position (G): (3,3)").
    \item A list of all obstacle coordinates (e.g., ``Bomb Locations (\#): [(0,2), (1,3)]").
    \item Explicit task rules, including avoiding obstacles, staying within boundaries, making adjacent moves (North, South, East, West), and not revisiting cells.
\end{itemize}
Two prompting strategies were employed for text-only models, using predefined templates for system and user messages:
\begin{enumerate}
    \item \textbf{Zero-shot (ZS):} The LLM was provided with a system prompt outlining its role as an expert spatial planner and the required output format. This was followed by a user prompt containing the specific textual grid description and the direct task to generate the path.
    \item \textbf{Few-shot (FS):} The system prompt was identical to the zero-shot condition. The user prompt, however, first presented one complete example consisting of a grid description, the task query, and the corresponding correct path solution, before presenting the actual problem instance to be solved.
\end{enumerate}

\subsubsection{Experiment 2: Multimodal LLM Evaluation}
\label{subsubsec:multimodal_eval}
Multimodal LLMs were evaluated using a specific prompting strategy designed to leverage both visual and textual input streams.
\begin{enumerate}
    \item \textbf{Multimodal Input and Prompting Strategy:}
    \begin{itemize}
        \item \textit{Input:} The LLM received the $512 \times 512$ PNG image of the grid (showing `S' and `G' locations, with obstacles as non-white cells) along with the complete textual description identical to that used for text-only models. This textual input included grid size, `S'/`G' coordinates, all obstacle locations, the optimal path length, and explicit task rules, consistent with the information provided in text-only experiments.
        \item \textit{Prompting Strategy:} The system prompt guided the LLM to perform a two-step reasoning process before generating the path:
        \begin{enumerate}
            \item \textbf{Visual Confirmation of Obstacles:} The LLM was first instructed to explicitly compare the obstacle locations listed in the textual description against the visual obstacles (non-white cells, excluding `S' and `G') in the provided image. It was asked to state whether these two sources of obstacle information were consistent.
            \item \textbf{Path Planning:} Subsequently, the LLM was tasked to generate the path trajectory based on integrating information from \textit{both} the textual description (with particular emphasis on the confirmed list of obstacle locations) and the visual context of the grid image.
        \end{enumerate}
        This structured prompting approach is designed to encourage the LLM to actively correlate information from both modalities and perform an explicit reasoning step prior to path planning.
    \end{itemize}
\end{enumerate}

\subsection{Trajectory Parsing and Evaluation Metrics}
The textual response from the LLM, expected to contain a Python list of coordinate tuples representing the path, was parsed to extract the trajectory. The extracted trajectory was then rigorously validated against the ground truth for the given grid problem through a comprehensive validation process. The validation criteria ensure the trajectory:
\begin{itemize}
    \item Starts at the designated `S' position and correctly ends at the `G' position.
    \item Consists exclusively of valid, adjacent moves (North, South, East, or West between consecutive coordinates).
    \item Remains entirely within the defined grid boundaries.
    \item Does not pass through any cells designated as obstacle locations.
    \item Does not revisit any cell along its length (a strictly enforced rule).
\end{itemize}

\begin{figure*}[htbp] 
  \centering
  \includegraphics[width=\textwidth]{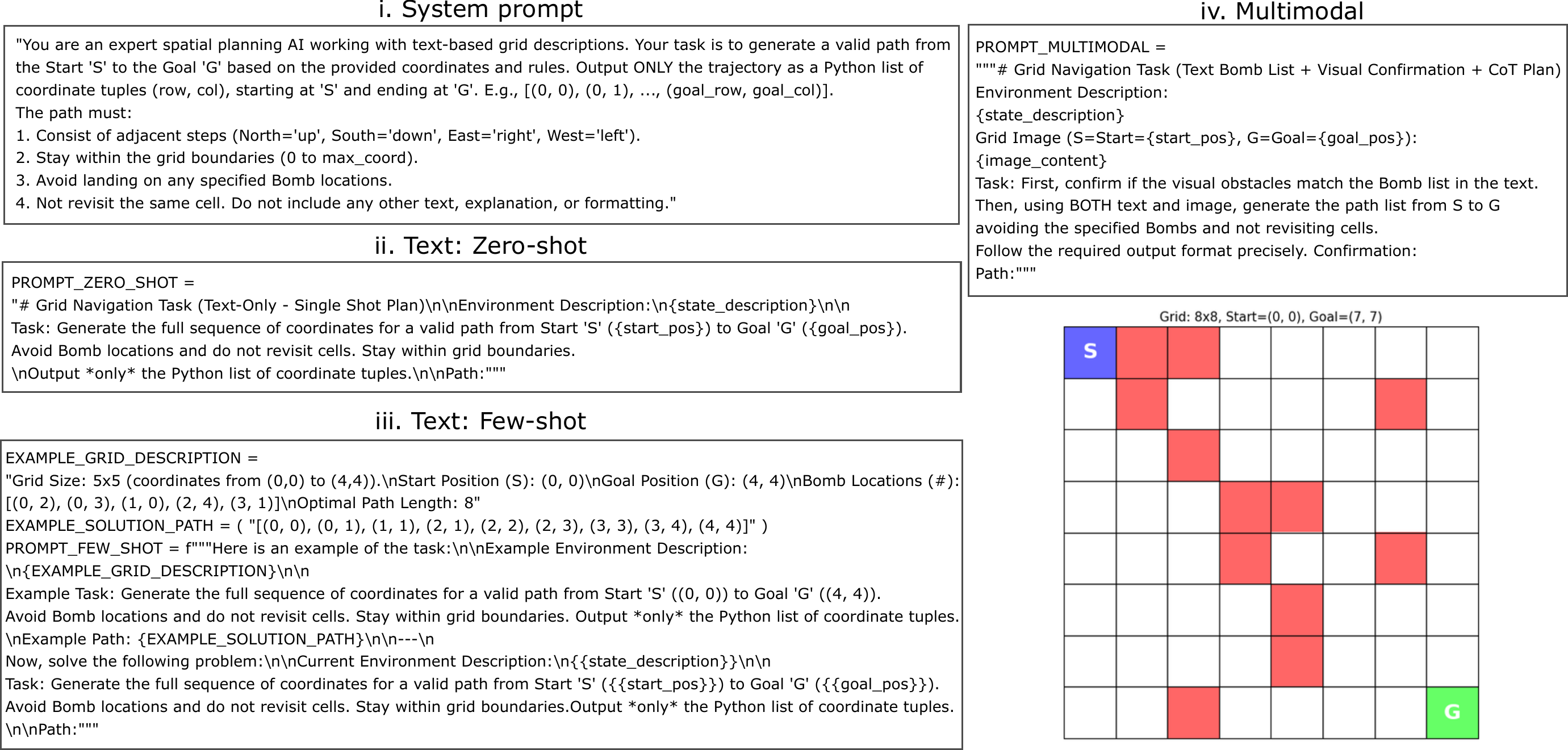} 
  \caption{Example prompt structures for LLM evaluation: \textbf{i.} System prompt (common for text-only tasks) outlining the AI's role and output format. \textbf{ii.} User prompt for text-only zero-shot evaluation, providing the grid description and task. \textbf{iii.} User prompt for text-only few-shot evaluation, including an example problem-solution pair before the target problem. \textbf{iv.} User prompt for multimodal evaluation, providing textual grid description, requesting visual confirmation of obstacles against the text, and then path generation using both text and image. Example of $8 \times 8$ grid visualization provided to multimodal models, showing Start (`S') and Goal (`G') positions.}
  \label{fig:1_revised} 
\end{figure*}

\section{Experimental Results}
\label{sec:results}

This section presents the quantitative results of our comparative benchmark, evaluating the spatial reasoning and path generation performance of selected Large Language Models (LLMs). We focus on Success Rate, Optimality Rate, Average Path Length Ratio, and Average Suboptimality Gap.

\subsection{Performance Metrics}
\label{subsec:metrics}
To evaluate model performance, we use the following metrics. Let $N$ be the total number of trials for a given condition. Let $V$ be the set of valid trials, and $|V|$ their count. Let $O$ be the set of valid and optimal trials, with $|O|$ their count. For a trial $i$, $L_{gen,i}$ is the generated path length, and $L_{opt,i}$ is the optimal path length.

\textbf{Success Rate:} Percentage of trials with a valid path.
$$ \text{Success Rate} = \frac{|V|}{N} \times 100\% $$

\textbf{Optimality Rate:} Percentage of \textit{valid} trials with an optimal path.
$$ \text{Optimality Rate} = \begin{cases} \frac{|O|}{|V|} \times 100\% & \text{if } |V| > 0 \\ 0\% & \text{if } |V| = 0 \end{cases} $$

\textbf{Average Path Length Ratio:} Ratio of generated to optimal path length for valid paths where $L_{opt,i} > 0$.
$$ \text{Avg. Path Length Ratio} = \frac{1}{|V'|} \sum_{i \in V'} \frac{L_{gen,i}}{L_{opt,i}} $$
where $V' = \{i \in V \mid L_{opt,i} > 0\}$. If $|V'|=0$, reported as 0.

\textbf{Average Suboptimality Gap:} Average extra steps for valid, non-optimal paths.
$$ \text{Avg. Suboptimality Gap} = \frac{1}{|V_{subopt}|} \sum_{i \in V_{subopt}} (L_{gen,i} - L_{opt,i}) $$
where $V_{subopt} = \{i \in V \mid L_{gen,i} > L_{opt,i}\}$. If $|V_{subopt}|=0$, the gap is 0.

\subsection{Overall Performance on Grid Path Generation}
\label{subsec:overall_performance}
 Success rates over all evaluated 15 LLMs were generally modest, underscoring the task's difficulty, especially on larger grids, which challenge scalable planning and precise constraint adherence.

\subsection{Impact of Prompting Strategies and Modality }
\label{subsec:strategy_modality_impact_8x8}

\begin{figure}[ht]
    \centering
    \includegraphics[width=0.45\textwidth]{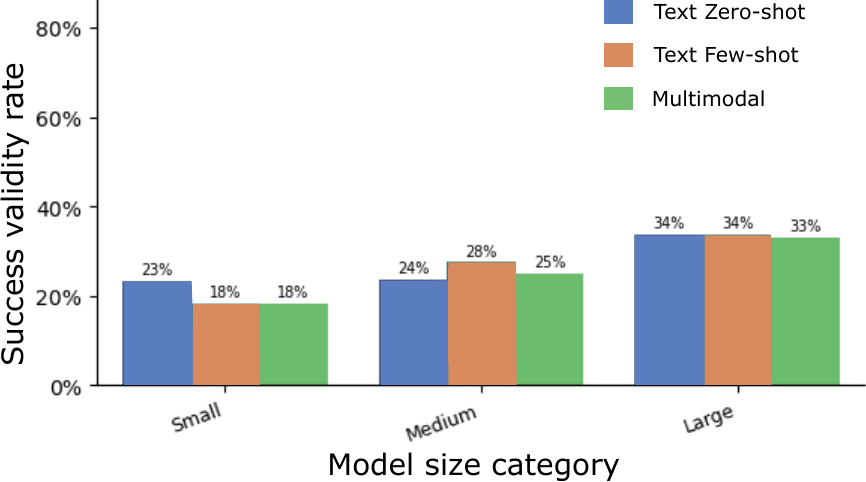}
    \caption{Average Success Rate by model size ($8 \times 8$ grids), comparing Text ZS, Text FS, and Multimodal prompt. }
    \label{fig:2}
\end{figure}

\textbf{Text-Only Models:} Figure \ref{fig:2} shows FS prompting slightly improved average Success Rates for Small (FS 24\% vs. ZS 23\%) and Medium (FS 25\% vs. ZS 20\%) models. For Large models, ZS (34\%) averaged higher than FS (28\%). This suggests that for larger models, a single few-shot example might not consistently generalize well or could offer less benefit if the model's zero-shot capabilities are already relatively strong for the core task understanding. Individual model results (Figures \ref{fig:3}, \ref{fig:4}, \ref{fig:5}) confirm this variance; for example, DeepSeek V3 (Large) improved with FS (60\% from 40\% ZS), while Llama 4 Maverick (Large) did not. Valid paths from text-only strategies were generally of high quality on $8 \times 8$ grids, with optimality rates often near 100\% and path length ratios close to 1.00 (Tables \ref{tab:2}, \ref{tab:3}).

\textbf{Multimodal Models:} This strategy yielded average Success Rates of 18\% (Small), 24\% (Medium), and 33\% (Large) on $8 \times 8$ grids (Figure \ref{fig:2}). Some Large models like Claude 3.7 Sonnet (55\%) performed well (Figure \ref{fig:5}). Successful multimodal paths demonstrated excellent quality, often being perfectly optimal (Tables \ref{tab:2}, \ref{tab:3}). This indicates effective information integration when successful, but achieving that initial success remains a hurdle.

\textbf{Comparative Analysis:} No single strategy was universally superior in Success Rate on $8 \times 8$ grids (Figure \ref{fig:2}). FS was competitive for Small/Medium models, while ZS and Multimodal led for Large models. The richer context from multimodal input did not always translate to higher success than focused text-only approaches, possibly due to the complexities of cross-modal grounding. Path quality metrics (Tables \ref{tab:2}-\ref{tab:4}) show that Small models were nearly perfect when valid. Medium models saw Text FS leading in optimality. For Large models, Multimodal had slightly better average optimality rates, though overall path quality differences among strategies were minor for successful paths.

\begin{table}[htb]
\caption{Average Path Length Ratio by model size and strategy for $8 \times 8$ grids. Values closer to 1.00 are better.}
\begin{center}
\label{tab:2}
\begin{tabular}{|c|c|c|c|}\hline
~ & ~Text-Zero~ & ~Text-Few~ & ~Multimodal~ \\\hline
Small & 1.000 & 1.000 & 1.000 \\\hline
Medium & 1.012 & 1.000 & 1.014 \\\hline
Large & 1.045 & 1.030 & 1.051 \\\hline
\end{tabular}
\end{center}
\end{table}

\begin{table}[htb]
\caption{Average Optimality Rate (\%) by model size and strategy for $8 \times 8$ grids.}
\begin{center}
\label{tab:3}
\begin{tabular}{|c|c|c|c|}\hline
~ & ~Text-Zero~ & ~Text-Few~ & ~Multimodal~ \\\hline
Small & 100\% & 100\% & 100\% \\\hline
Medium & 96\% & 100\% & 95\% \\\hline
Large & 81\% & 82\% & 80\% \\\hline 
\end{tabular}
\end{center}
\end{table}

\begin{table}[htb]
\caption{Average Suboptimality Gap (extra steps) by model size and strategy for $8 \times 8$ grids. Lower is better.}
\begin{center}
\label{tab:4}
\begin{tabular}{|c|c|c|c|}\hline
~ & ~Text-Zero~ & ~Text-Few~ & ~Multimodal~ \\\hline
Small & 0.00 & 0.00 & 0.00 \\\hline
Medium & 1.00 & 0.00 & 1.00 \\\hline
Large & 2.42 & 1.88 & 2.21 \\\hline
\end{tabular}
\end{center}
\end{table}

\subsection{Performance Variation Across Model Size }
\label{subsec:model_size_variation_8x8}
\begin{figure}[ht]
    \centering
    \includegraphics[width=0.45\textwidth]{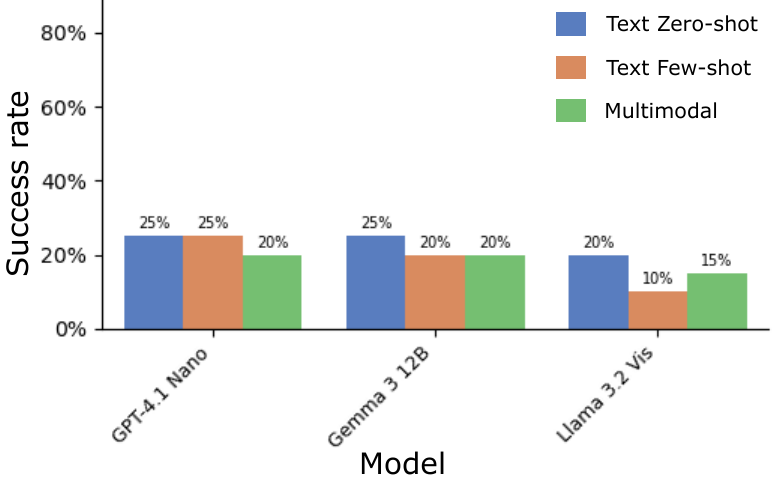}
    \caption{Success Rate for Small size models ($8 \times 8$ grids). }
    \label{fig:3}
\end{figure}

\begin{figure}[ht]
    \centering
    \includegraphics[width=0.45\textwidth]{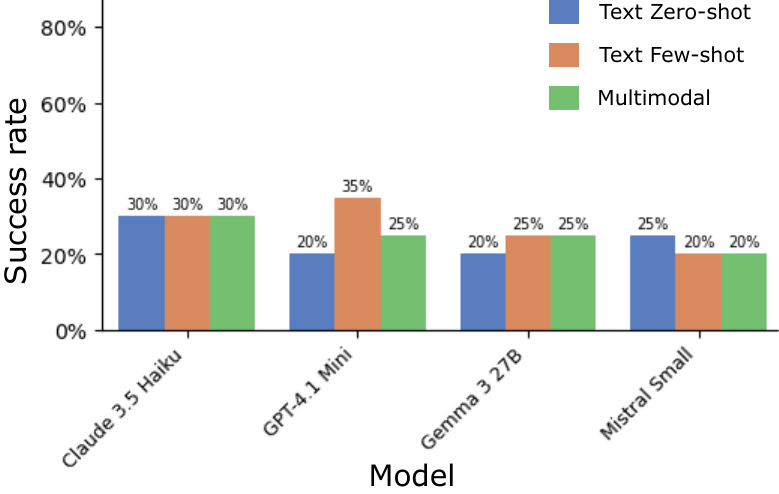}
    \caption{Success Rate for Medium size models ($8 \times 8$ grids). }
    \label{fig:4}
\end{figure}

\begin{figure}[ht]
    \centering
    \includegraphics[width=0.45\textwidth]{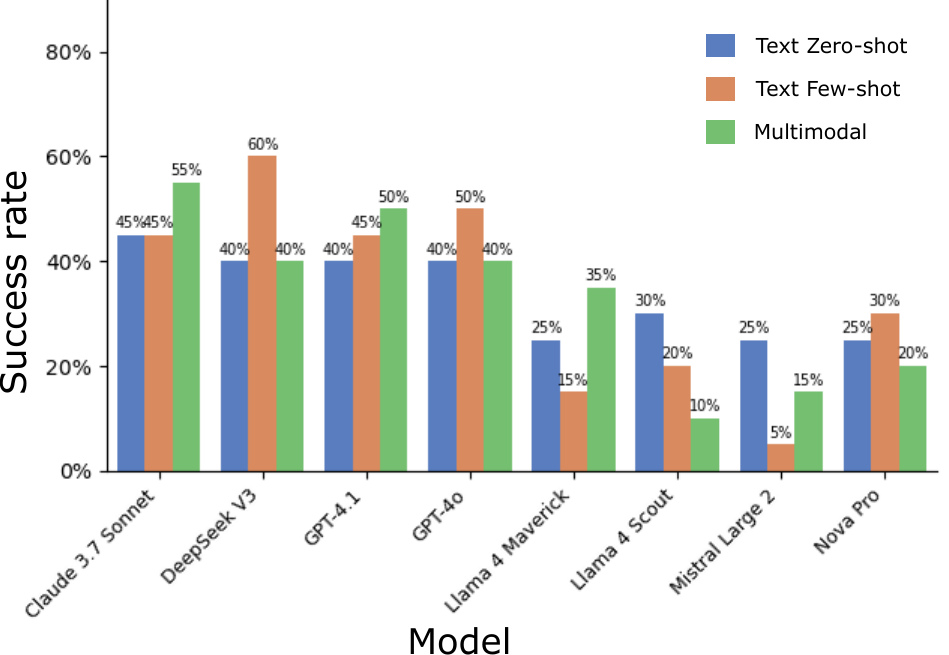}
    \caption{Success Rate for Large size models ($8 \times 8$ grids). }
    \label{fig:5}
\end{figure}

Individual model Success Rates on $8 \times 8$ grids are shown in Figures \ref{fig:3} (Small), \ref{fig:4} (Medium), and \ref{fig:5} (Large).

Larger models generally achieved higher peak Success Rates (e.g., up to 55-60\% for some Large models, Figure \ref{fig:5}), while Medium models performed in the 20-35\% range (Figure \ref{fig:4}), and Small models typically below 25\% (Figure \ref{fig:3}). This trend, also seen in the aggregated Figure \ref{fig:2}, suggests model scale aids in handling the task's rule complexity and multi-step reasoning. However, significant variance within the Large category indicates scale is not the sole factor; architecture and training likely play crucial roles.

\subsection{Influence of Grid Complexity:}
\label{subsec:grid_complexity_influence}
\begin{figure}[htbp]
    \centering
    \includegraphics[width=0.3\textwidth]{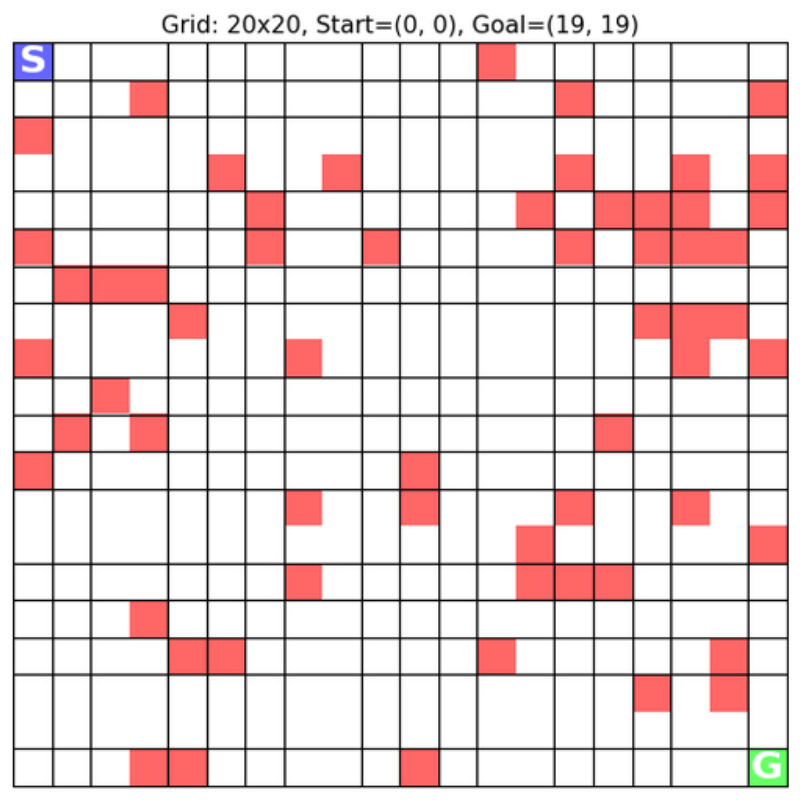}
    \caption{Example of a $20 \times 20$ grid environment.}
    \label{fig:example_20x20_grid_render}
\end{figure}

\begin{figure}[htbp]
    \centering
    \includegraphics[width=0.45\textwidth]{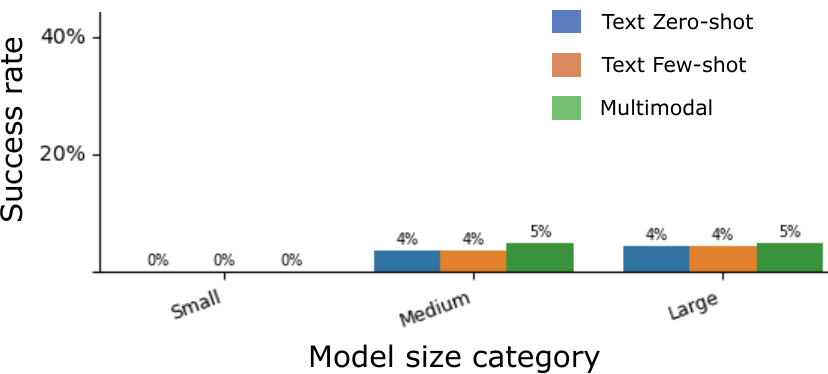}
    \caption{Average Success Rate by model size ($20 \times 20$ grids). }
    \label{fig:avg_validity_20x20_grids}
\end{figure}

Performance degraded sharply on $20 \times 20$ grids (example in Figure \ref{fig:example_20x20_grid_render}) compared to $8 \times 8$ grids. Average Success Rates (Figure \ref{fig:avg_validity_20x20_grids}) plummeted to nearly 0\% for Small models, 0-4\% for Medium, and only 4-5\% for Large models.

This highlights a critical scalability challenge. The quadratically increased state space, longer paths, and complex obstacle configurations in $20 \times 20$ grids likely exceed the capacity of current models for sustained, precise planning, and constraint adherence. 
The path quality for the rare successful $20 \times 20$ paths also worsened, with higher path length ratios (e.g., 1.3-1.4 for some large models) and larger suboptimality gaps (10-30+ extra steps). This suggests that maintaining global optimality or even near-optimality is exceptionally difficult in larger environments, as models may default to locally satisfactory but globally inefficient path segments.

\section{Conclusion}
\label{sec:conclusion}
This study systematically benchmarked the performance of inherently multimodal Large Language Models on grid-based path planning tasks, simulating simplified robotic planning scenarios to assess the utility of visual input alongside textual descriptions. Our findings indicate that while contemporary multimodal LLMs can achieve moderate success rates in generating valid paths on simpler grids, with some models benefiting from visual input or specific prompting strategies such as few-shot learning, their performance significantly degrades on more complex grids. This highlights a critical scalability challenge. Although larger models tended to exhibit higher average success rates, the benefit of visual modality was not universally dominant over well-structured textual input for these multimodal systems, and the path quality for successful attempts was often high regardless of the input modality on simpler grids. These results underscore current limitations in robust spatial reasoning, precise adherence to constraints over extended planning horizons, and effective, scalable integration of multimodal information for these models. Future research should therefore prioritize enhancing these aspects to improve the reliability and applicability of LLMs in complex robotic path planning and autonomous navigation.


\begin{thebibliography}{9}

\bibitem{ref1_old}
L. Wang, C. Ma, X. Feng, Z. Zhang, H. Yang, J. Zhang, Z. Chen et al., ``A survey on large language model based autonomous agents'', {\it Frontiers of Computer Science}, Vol. 18, No. 6, pp. 186345, 2024.

\bibitem{ref2_old}
J. Wang, E. Shi, H. Hu, C. Ma, Y. Liu, X. Wang, Y. Yao, X. Liu, B. Ge, and S. Zhang., ``Large language models for robotics: Opportunities, challenges, and perspectives'', {\it Journal of Automation and Intelligence}, Vol. 4, No. 1, pp. 52--64, 2025.

\bibitem{ref1}
J. Colan, A. Davila, Y. Yamada, and Y. Hasegawa, ``Human-Robot collaboration in surgery: Advances and challenges towards autonomous surgical assistants'', {\it IEEE International Conference on Robot and Human Interactive Communication (ROMAN)}, 2025.

\bibitem{ref2}
W. Huang, P. Abbeel, D. Pathak, and I. Mordatch., ``Language models as zero-shot planners: Extracting actionable knowledge for embodied agents'', {\it International conference on machine learning}, pp. 9118--9147, 2022.

\bibitem{ref3}
B. Ichter, A. Brohan, Y. Chebotar, et al., ``Do As I Can, Not As I Say: Grounding Language in Robotic Affordances'', {\it Proceedings of The 6th Conference on Robot Learning}, Vol. 205, pp. 287--318, 2023.

\bibitem{ref4}
I. Singh, V. Blukis, A. Mousavian, A. Goyal, D. Xu, J. Tremblay, D. Fox, J. Thomason, and A. Garg., ``Progprompt: Generating situated robot task plans using large language models'', {\it 2023 IEEE International Conference on Robotics and Automation (ICRA)}, pp. 11523--11530, 2023.

\bibitem{ref5}
A. Davila, J. Colan, and Y. Hasegawa, ``Voice control interface for surgical robot assistants'', {\it 2024 International Symposium on Micro-NanoMehatronics and Human Science (MHS)}, pp. 1--5, 2024.

\bibitem{ref6}
A. Davila, J. Colan, and Y. Hasegawa, ``LLM-based ambiguity detection in natural language instructions for collaborative surgical robots'', {\it 2025 IEEE International Conference on Robot and Human Interactive Communication (ROMAN)}, 2025.

\bibitem{ref7}
K. Fozilov, J. Colan, K. Sekiyama, and Y. Hasegawa, ``Toward autonomous robotic minimally invasive surgery: A hybrid framework combining task-motion planning and dynamic behavior trees'', {\it IEEE Access}, Vol. 11, pp. 91206--91224, 2023.


\bibitem{ref8}
S. Chen, Z. Wu, K. Zhang, C. Li, B. Zhang, F. Ma, F. R. Yu, and Q. Li., ``Exploring embodied multimodal large models: Development, datasets, and future directions'', {\it Information Fusion}, pp. 103198, 2025.

\bibitem{ref7_old}
X. Han, S. Chen, Z. Fu, Z. Feng, L. Fan, D. An, C. Wang, L. Guo, W. Meng, X. Zhang, and R. Xu., ``Multimodal Fusion and Vision-Language Models: A Survey for Robot Vision'', {\it arXiv:2504.02477}, 2025.

\bibitem{ref9}
D. Driess, F. Xia, M. S. M. Sajjadi et al., ``PaLM-E: An Embodied Multimodal Language Model'', {\it arXiv:2303.03378}, 2023

\bibitem{ref1_old0}
A. Brohan, N. Brown, J. Carbajal, Y. Chebotar, X. Chen, K. Choromanski, T. Ding et al., ``Rt-2: Vision-language-action models transfer web knowledge to robotic control'', {\it arXiv:2307.15818}, 2023.

\bibitem{ref1_old1}
Y. Jiang, A. Gupta, Z. Zhang, G. Wang, Y. Dou, Y. Chen, L. Fei-Fei, A. Anandkumar, Y. Zhu, and L. Fan., ``VIMA: General Robot Manipulation with Multimodal Prompts'', {\it arXiv:2210.03094}, 2023.

\bibitem{ref1_old2}
S. Meng, Y. Wang, C.-F. Yang, N. Peng, and K.-W. Chang., ``LLM-A*: Large Language Model Enhanced Incremental Heuristic Search on Path Planning'', {\it arXiv:2407.02511}, 2025.

\bibitem{ref1_old3}
L. Xiao and T. Yamasaki., ``LLM-Advisor: An LLM Benchmark for Cost-efficient Path Planning across Multiple Terrains'', {\it arXiv:2503.01236}, 2025.

\bibitem{ref1_old4}
P. Doma, A. Arab, and X. Xiao., ``LLM-Enhanced Path Planning: Safe and Efficient Autonomous Navigation with Instructional Inputs'', {\it arXiv:2412.02655}, 2024.

\bibitem{ref1_old5}
S. Song, S. Kodagoda, A. Gunatilake, M. G. Carmichael, K. Thiyagarajan, and J. Martin., ``Guide-LLM: An Embodied LLM Agent and Text-Based Topological Map for Robotic Guidance of People with Visual Impairments'', {\it arXiv:2410.20666}, 2025

\bibitem{ref1_old6}
N. Martorell., ``From Text to Space: Mapping Abstract Spatial Models in LLMs during a Grid-World Navigation Task'', {\it arXiv:2502.16690}, 2025.

\bibitem{ref1_old7}
H. Zhang, C. Li, W. Wu, S. Mao, Y. Xia, I. Vulić, Z. Zhang, L. Wang, T. Tan, and F. Wei., ``A Call for New Recipes to Enhance Spatial Reasoning in MLLMs'', {\it arXiv:2504.15037}, 2025.

\bibitem{ref1_old8}
Y.-H. Shen, C.-Y. Wu, Y.-R. Yang, Y.-L. Tai, and Y.-T. Chen., ``Mitigating Cross-Modal Distraction and Ensuring Geometric Feasibility via Affordance-Guided, Self-Consistent MLLMs for Food Preparation Task Planning'', {\it arXiv:2503.13055}, 2025.

\bibitem{ref1_old9}
M. Aghzal, E. Plaku, and Z. Yao., ``Can Large Language Models be Good Path Planners? A Benchmark and Investigation on Spatial-temporal Reasoning'', {\it arXiv:2310.03249}, 2025.

\bibitem{ref2_old0}
R. Yang, H. Chen, J. Zhang, et al., ``EmbodiedBench: Comprehensive Benchmarking Multi-modal Large Language Models for Vision-Driven Embodied Agents'', {\it arXiv:2502.09560}, 2025

\end{thebibliography}
\end{document}